\documentclass[journal]{IEEEtran}
\usepackage{lettrine}
\usepackage{epsf}
\usepackage{bm}
\usepackage{algorithm}
\usepackage{tabularx}
\usepackage[noend]{algpseudocode}
\usepackage{dcolumn}
\usepackage{graphicx}
\usepackage{setspace}
\usepackage[export]{adjustbox}
\usepackage{subcaption}
\usepackage{subfig}
\usepackage{multicol}
\usepackage{slashbox}
\usepackage{array}
\usepackage{mathtools}

\title{New Methods of Enhancing Prediction Accuracy in Linear Models with Missing Data}

\author{Mohammad Amin Fakharian, Ashkan Esmaeili, and Farokh Marvasti\\ Department of EE, Sharif University of Technology\\
	Advanced Communications Research Institute (ACRI)\\  fakharian\_ma@ee.sharif.edu, ashkan.esmaeili@ee.sharif.edu, marvasti@sharif.edu}

\begin{document}
\pagenumbering{gobble}
\maketitle
\begin{abstract}
In this paper, prediction for linear systems with missing information is investigated. New methods are introduced to improve the Mean Squared Error (MSE) on the test set in comparison to state-of-the-art methods, through appropriate tuning of Bias-Variance trade-off.  First, the use of proposed Soft Weighted Prediction (SWP) algorithm and its efficacy are depicted and compared to previous works for non-missing scenarios. The algorithm is then modified and optimized for missing scenarios. It is shown that controlled over-fitting by suggested algorithms will improve prediction accuracy in various cases. Simulation results approve our heuristics in enhancing the prediction accuracy.
\end{abstract}


\section{Introduction}\label{intro}
\lettrine{R}{ecently}, there has been a growing interest in enhancing prediction accuracy in Machine Learning (ML). Although previous studies indicate that clustering may improve accuracy \cite{Clustering}, training set shrinkage and data ignorance would be the penalties since it assigns hard weights to the subjects (i.e. each member has a weight parameter $w=\{0,1\}$). Mentioned penalties result in uncontrolled over-fitting in various cases. In this paper, a novel method of classification is presented. We call this method Soft Weighted Prediction (\rm{SWP}), which weighs each cluster obtained from training set (possibly each training example if they form a cluster themselves) based on its Euclidean distance from each test set subject.\\
Missing information has been gaining importance quite recently due to wide vision of applications it accompanies in practice. Although several methods of clustering for such scenarios are developed and introduced, none of them focus on missing information patterns. An innovative method of clustering without matrix completion is introduced in this paper. Soft Constrained clustering (SCOP) concept, introduced by Kiri Wagstaff \cite{KSCOP}, is a prototypical useful tool in the algorithm. The solution we suggest is compared to imputation algorithms, which are the most common approaches in dealing with missing information.\\
Missing parameters in medical datasets for instance, caused by data loss or idleness could be considered as a practical paradigm of inducing data loss in the structure of prediction problem. Obviously, in such cases missing values are not randomly distributed, e.g. patients suffering from the same disease, are more likely to be recorded with the same factors and symptoms. Thus, patients with similar missing factors, tend to be clustered together and have tendency to be reported with correlated medical diagnosis. This lack of similar recorded parameters (jointly missing parameters for subjects) is supposed to be a constraint i n clustering.

\section{Model Assumptions}\label{Problem Formulation}
In matrix representation, linear models are depicted as follows:
\begin{equation}
Y = X\beta + \varepsilon,
\end{equation}
where $ \varepsilon \sim N(\mu,\sigma) $, 
$\mathbf{X}$ is the data matrix consisting of subjects parameters in the true model. However, in practice, we partially observe the entries of $\mathbf{X}$, and it is assumed that the matrix provided is obtained by putting a mask on the original data matrix. The mask contains zeros on the entries which are missing or lost, i.e. we have access to a data matrix $\mathbf{\tilde{X}} = \mathbf{X}\bigodot \mathbf{M}$
, where $\mathbf{M}$ is the oracle mask, $Y$ is the observed measurement vector, and $\beta$ is parameters (weights) coefficients. 
\subsection{Bias-Variance Trade-Off}
As the following equation states, \rm{$MSE$} consists of three terms. It is supposed that the noise variance is fixed; therefore, optimal prediction is achieved through balancing variance and bias terms in the decomposition provided in \ref{eq:2}.
\begin{equation} \label{eq:2}
E[(y - \hat {f} (x))^2] = (Bias[\hat{f}(x)])^2 +Var[\hat{f}(x)] + \sigma^2,
\end{equation}
where
\begin{equation}
Bias[\hat{f}(x)]=E[\hat {f} (x)]-f(x),
\end{equation}
and
\begin{equation}
Var[\hat{f}(x)]=E[(\hat {f} (x)-E[\hat{f}(x)])^2].
\end{equation}
\subsection{Mathematical Approaches in Extracting the True Model }
Coefficients vector $\beta$ could be estimated knowing $\mathbf{X}$ and $Y$ as $b$. There are several regularization methods based on assumed constraints on vector $\beta$  such as sparsity, to find the estimator $b$ as it is not unique in many cases. However, our main concern is superior prediction of vector $Y$, not the coefficient. As Lasso constrains desired over-fitting, the Least-Square (LS) solution to the problem is used in the algorithms.\\

\subsubsection{Lasso Solution}
Supposing $\beta$ as a sparse vector, desired $b$ will be obtained satisfying condition \ref{eq:3}.

\begin{equation} \label{eq:3}
\min_{b} \frac{1}{2} ||Y-Xb||^2_2 + \lambda||b||_1,
\end{equation}
where parameter $\lambda$ controls the sparsity rate of coefficient $\beta$ which is equivalent to balancing the trade-off.\\
Supposing $\lambda=0$, our problem model turns into unconstrained problem, or ordinary least square. As $\lambda$ approaches zero this solution will have less bias and more variance error terms. Thus, this concept is a data dependent (training set) solution. As a result, test and train variation will lead to an inferior estimation and larger MSE. Further, as $\lambda$ approaches $\infty$, $b$ will be constrained to be sparse. Thus, training set variation effect decreases and estimator data dependency will be omitted.\\

\subsubsection{Least-Square Solution}

The $LS$ solution is a particular case of $LASSO$ ($\lambda=0$). Solution to the problem is a vector $b$ estimating coefficient $\beta$. The normal equations are as follows:
$$
(X^TX)b=X^T Y
$$
Solving for $b$,
\begin{equation} 
b = (X^T X)^{-1} X^T Y
\end{equation}

Let $Y=X\beta$, adding noise $\varepsilon \sim N(0,1)$ to  the $LS$, the solution of the problem will be:

\begin{equation}
b = (X^TX)^{-1}X^TY+(X^TX)^{-1}X^T \varepsilon
\end{equation}
\begin{equation}
b = \beta+(X^TX)^{-1}X^T \varepsilon
\end{equation}
The expected value is:
\begin{equation}
E[b] = E[\beta]+(X^TX)^{-1}X^T E[\varepsilon]
\end{equation}
Knowing that $E[\varepsilon]=0$,
\begin{equation}
E[b] = \beta
\end{equation}
Thus, $LS$ is the desired unbiased solution to the problem.

\subsection{Overfitting}
Overfitting occurs in test and training set variation cases. This error could be controlled by constraining the training set based on its similarity to each test example. This constraining could be done by either soft or hard weighting methods. In hard weighting algorithms training set would be shrunken to the most similar members to test example, such as clustering. On the other hand, soft weighting method prevents such data losses by a weighting mask based on similarities. Although $SWP$ methods may cause accuracy reduction for estimator $b$ specifically in sparse cases, more accurate $Y$ estimation will be obtained. Specific estimator $b$ is calculated for each test member based on its distance from $X$, which is not necessarily a good estimation of $\beta$, but more accurate prediction for $Y$. As overfitting is controlled (by similarity) and is satisfactory in such scenarios. Therefore, overfitted $b$ is not our main concern, e.g. introduced clustering algorithm, segments $X$ and allocates each test set example, a cluster based on its Euclidean distance from its centroid. Thus, estimator $b$ is trained by specific members, which results in increase of variance and reduction in bias term of predicted $Y$ error. By increasing the number of clusters, overfitting and increase in variance term error will be observed.

\section{Proposed Algorithm}
Clustering as a method of tuning variance-bias trade-off has been studied in the literature as in \cite{Clustering}. Although simulations depicted enhancement of prediction responses in some cases, hard clustering results in uncontrolled overfitting and data loss. The efficiency of hard clustering in comparison to suggested algorithms is more deeply investigated.\\ 
K-mapping is one of the methods trying to optimize Bias-Variance trade-off. The error expression is in this case:
\begin{equation}
E[(y-\hat{f}(x))^2]=(f(x)-\frac{1}{k}\sum_{i=1}^{k}f(N_i(x)))^2+ \frac{\sigma^2}{k}+\sigma^2
\end{equation}
Supposing $k$ nearest neighbors are chosen from the training set, bias which is the first term, has a monotonous rise as $k$ increases. On the other hand, variance reduces at the same time.\\
Although variance minimization leads to worse interpolation of training set, dependent on its answer $Y$, it removes data dependency. Bias minimization has the reverse effect, i.e. although estimator $b$ leads to the best $Y$ calculation dependent to the specific training set $X$, vector $b$ itself has larger $MSE$ to the real coefficient coefficient $\beta$. Obviously in such cases if test matrix does not fit in any of the clusters, the estimated $Y$ will face a larger error (large variance and small bias).

As K-means algorithm with squared Euclidean distance parameter is chosen for k-mapping, in order to specify appropriate cluster for each individual, the centroids of clusters are kept in a matrix \rm{$C$}. Thus, Minimum n-dimensional distance of test set example to each row of matrix $C$, leads to the appropriate cluster. Following the $LS$ fitting solution, the predicted $b$ is found. Multiplying test and estimator $b$, results in predicted $Y$ matrix. As the number of clusters ($k$) increase, members of each cluster will decrease. Although this will lead to lower bias, variance term of error will increase. If test varies from training set, Estimated $Y$ accuracy will be greatly depressed.\\

Proposed solution to the problem is comprised of assigning to each training set subject, a specific weight based on its similarity to the test. This filter is set to be an exponential function of distance. $\mathbf{W}$ is a $m \times 1$ matrix (filter) containing normalized n-dimensional distance between test and each training set subject. Parameter $w$ controls the strength of filtering. As it approaches infinity, filter approaches one (no filtering).\\

\begin{algorithm}[t]
\begin{spacing}{1.5}
\textbf{Input:} Training set $X_{train}$, Training response vector $Y_{train}$, Test set $X_{test}$, Weight tuning parameter $w$.\\
\textbf{Output:}  Test set response vector $Y_{test}$.
\begin{algorithmic}[1]
\caption{SWP}
\Function{SWP}{$X_{train},Y_{train},X_{test},w$}
\ForAll{$X_{test}(i,:)$}
\State{$data_{new}=X_test{i,:}$}
\State $diff(j) := ||data_{new}-X_{train}(j,:)||^2_2$
\State $diff \gets \frac{diff}{min(diff)}$
\State $W := diag(e^{\frac{-diff}{2^w}})$
\State $b \gets (X_{train}^TWX_{train})^{-1}X_{train}^TWY_{train}$
\State $Y_{test}(i,:) \gets data_{new} \times b$ 
\EndFor
\State \textbf{end for}
\State \Return $Y_{test}$
\EndFunction
\State \textbf{end function}
\end{algorithmic}
\end{spacing}
\end{algorithm}

Obviously, all sub-figures of Fig. \ref{fig:1} in \ref{SWP_nmissing} depict the same behavior which caused by Bias-Variance tradeoff.\\

\section{Missing Values}
\label{missing values}
Introduced methods are dependent on data matrix  (training set). Considering missing values, clustering wouldn't be possible (by k-means). Therefore SWP algorithm requires a new definition of similarity too.

\subsection{Imputation Methods}\label{Imputation}
\subsubsection{Soft Impute \cite{SVD}}
In this method, $Z$ is considered as a low-rank matrix. As $rank(Z)$ is a non-convex function, relaxation could be carried out by minimizing equivalent nuclear norm of $Z$. Finding matrix $Z$ which satisfies \ref{SI}, is desired.
\begin{equation}\label{SI}
||X-Z||_2^2~subject~to~||Z||_*\leq \tau
\end{equation}

\begin{equation}
\min_{Z} \frac{1}{2}||X-Z||_F^2+\lambda||Z||_*
\end{equation}
Soft-Thresholded SVD solution is:
$$
S_\lambda:=U(S-\lambda I)_{+}V^T
$$
Where $(S-\lambda I)_+$ is either positive or zero, otherwise.\\

To optimize the algorithm time complexity, the proposed idea is to initialize $Z$ from the mean estimation which results in more robustness in implementation. \\

\begin{figure*}
	\centering
	\includegraphics[trim={1.2cm 1.8cm 2cm 1.8cm},clip,width=\linewidth] {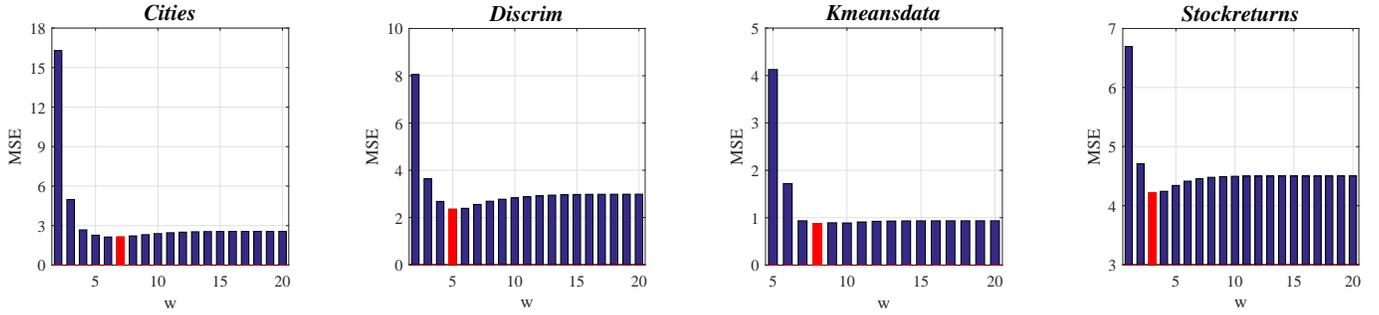}
	\caption{$MSE$ as a function of weight tuning parameter $w$.}
	\label{fig:1}
\end{figure*}

\subsection{Non-Impute Method}
Soft-Impute, an Imputation method, applies low-rank restriction on the recovered dataset. Data loss is an inevitable consequence of the solution, as linearly dependent features could be ignored in clustering.\\ Many recent studies have focused on clustering datasets containing missing informations. Most common suggested solutions offer modifications to clustering algorithms such as \textsc{Kmeans} and \textsc{FCM} which are illustrated in \cite{KSC} and \cite{FCM}, respectively. Although the main concern in such solutions are similarity of observed elements, it is worth noting that the same missing features represent a kind of resemblance in such scenarios. Balancing $n$-dimensional distance between observed data and missing features similarity by a weight tuning parameter leads to the desired clustering.\\

\subsubsection{Missing-SCOP}
We have chosen \rm{SCOP-KMEANS} Algorithm \cite{KSCOP} as a baseline for the development of missing values clustering. As the real model dictates, missing elements assume a role in clustering as a factor of similarity. By assuming missing mask similarity of each pair in training set as a constraint, our desire will be satisfied. Let matrix $S$ be an $m\times m$ matrix, which assigns a constraint $s=[-1,1]$ to each $(x_i,x_j)\in \mathbf{X\times X}$. Mentioned $s$ is set based on masks similarities and common observed features difference whose weights are tuned by a proportional tuning parameter $w$. As $s$ approaches -1, the constraint forces separation. On the other hand, when $s$ is 1, the two members of the pair must be clustered in the same group.\\

Replicative Kmeans algorithm is employed in centroid initialization due to local minimum solutions prevention.\\

\begin{algorithm}[t]
\begin{spacing}{1.5}
\textbf{Input:} Training set $X$, Number of Clusters $k$, Proportional Tuning Parameter $w$.\\
\textbf{Output:} Index vector $idx$, Centroids matrix $C$.
\begin{algorithmic}[1]
\caption{Missing-SCOP}
\label{Missing-SCOP}
\Function{Missing\_SCOP}{$X,k,w$}
\State $mask$ := not($X$==0)
\ForAll{i,j}
\If{i==j}
	\State continue
\EndIf
\State \textbf{end if}
\State $D_{miss}(i,j) := ||mask(x_i)-mask(x_j)||^2_2$
\State $co\_mask(i,j) := mask(i,:) \odot mask(j,:)$
\State $D_{dist}(i,j) := ||x_i-x_j||^2_2 \odot co\_mask(i,j)$
\State $D(i,j) := w \times D_{miss}(i,j) + (1-w) \times D_{dist}(i,j)$
\EndFor
\State \textbf{end for}
\State $S(i,j)=1 - 2\sqrt{\frac{D(i,j)}{\max{D(:)}}}$
\State $[idx,C] \gets$ SCOP\_KMEANS~\cite{KSCOP}~$(X,k,S)$
\EndFunction
\State \textbf{end function}
\end{algorithmic}
\end{spacing}
\end{algorithm}

\subsubsection{SWP via Missing-SCOP}
SWP algorithm consists of splitting the training set to one member clusters, and specifying each cluster a weight based on its distance to each individual. Another solution to the problem is soft clustering algorithms utilization to find the probability matrix $\mathbf{U}$ for the test example. Thus, weight matrix is a diagonal matrix in which members of same clusters have the same weights.\\
As the problem contains missing values, introduced Missing-SCOP algorithm is used to obtain more precise clustering in comparison to imputation methods.\\ 
Let $\mathbf{X}$ be the dataset matrix, divided to $m\times n$ train set $\mathbf{X_{train}}$ and $p\times n$ test set $\mathbf{X_{test}}$. Assuming $\mathbf{X_{train}}$ is clustered into $k$ sub-matrices by centroid matrix $\mathbf{C}$ and index vector $idx$, probability matrix $\mathbf{U}$ is defined as follows:

\begin{equation} \label{weight matrix}
U = 
\begin{bmatrix}
u_{11} & u_{12} & \cdots & u_{1k} \\
u_{21} & u_{22} & \cdots & u_{2k} \\
\vdots & \vdots & \ddots & \vdots \\
u_{p1} & u_{p2} & \cdots & u_{pk} \\
\end{bmatrix},
\end{equation}
where for each $i\in[1,p],~j\in[1,k]$
\begin{equation}
u_{ij}:=\frac{min\{u_{i1},u_{i2},...,u_{ik}\}}{||X_{test}(i,:)-C(j,:)||_2^2}
\end{equation}

Weight matrix $W$ in $SWP$ algorithm would be obtained by matrix $\mathbf{U}$, consequently. As $u_{ij}$ is a normalized factor of similarity between $i^{th}$ test set example and $j^{th}$ cluster centroid, vector $W_{clusters}$ and matrix $\mathbf{W}$ are defined for each $\mathbf{X_{test}}$ example in \ref{W_clusters_missing} and \ref{W_missing} respectively.\\

\begin{equation}
\label{W_clusters_missing}
W_{clusters}:=e^{\frac{-(\mathbf{U}(i,:)^{-1})}{2^w}},
\end{equation}
which is calculated for $i^{th}$ $\mathbf{X_{test}}$ example.
\begin{equation}
\label{W_missing}
\mathbf{W}:=diag\Big(W_{clusters}(j)\times(idx==j)\Big),
\end{equation}

where $j\in[1:k]$.\\

Weighted LS solution in the algorithm requires matrix completion which could be obtained by MIMAT \cite{esmaeili2016iterative} algorithm.\\

\section{Simulation Results}\label{Simulation Results}
\subsection{Datasets}\label{Models}
\subsubsection{Simulated Data}\label{Simulated Data}

As the real problems dictate, training set  and test set are random processes which consist of normally distributed random sequences (features).
Let  $X$ be an $m\times n$ random process consists of random variables  $X=\{X_1,X_2,...,X_n\}$ where $X_1,X_2,...,X_n$ are normally distributed with uniformly random parameters i.e. $X_i \sim N(\mu,\sigma)$.
As Law of Large Numbers ($LLN$) states, the average of the results obtained from a large number of trials should be close to the expected value, and will tend to become closer as more trials are performed. Due to data-dependency of the simulation results, our reported $MSE$s are averaged on 20 generated random data.

\subsubsection{Sample Data}\label{Sample Data}

Algorithms are also tested on following \textsc{Matlab} sample datasets:
$$cities,~discrim,~kmeansdata,~stockreturns$$

\subsubsection{Missing Mask}
Real cases depict significant and meaningful similarities in missing patterns of similar elements.\
Suggested missing mask consists of similar missing pattern for each cluster in $Dataset$ matrix. A Gaussian logic mask is added to this mask as expected in real world.\
Considering $m \times n$ dataset $\mathbf{X}$ clustered into $k$ sub-matrices each consisting of $n_1,n_2,...,n_k$ members based on index vector $idx$. The mentioned $m\times n$ logic mask is generated as described in \ref{eq:14}.

\begin{equation} \label{eq:14}
mask(idx==i,:):=ones(n_i,1)\times\Big(r\geq(r_{max}\times m_{rate})\Big),
\end{equation}

where $i=[1:k]$, $r_{max}=max(r(:))$, $m_{rate}$ is the missing rate and $r_{1\times n} \sim \mathbf{unif}$.

\subsection{No Missing Scenario}\label{nmissing case}
\subsubsection{SWP}\label{SWP_nmissing}
Algorithm is tested on datasets described in \ref{Models}. Results are respectively depicted in Fig. \ref{fig:1}. Although optimal tuning parameter $w$ varies from case to case, general behavior of the figures are the same.

\subsection{Missing Scenario}
Introduced methods dealing with missing elements of training set, are tested on mentioned datasets.\\

\subsubsection{Clustering}
\begin{figure}
	\centering
	\includegraphics[width=\linewidth] {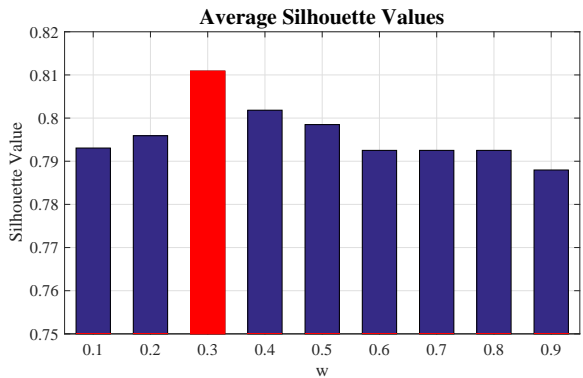}
	\caption{Averaged Silhouette Values as a function of weight tuning parameter $w$ tested on $kmeansdata$.}
	\label{fig:2}
\end{figure}

Our main concern in dealing with missing cases is clustering. Impute and Non-impute methods, introduced in Section \ref{missing values} are tested on datasets explained in \ref{Models}, which are masked by the mentioned method.\\
Silhouettes \cite{Silhouette} as a well-known method of clustering accuracy assessment is utilized. Simulation results are depicted in TABLE \ref{tab1} to compare and find the efficiency of each clustering algorithm.\\
Silhouette values of $kmeansdata$ as an appropriate dataset for clustering are depicted in fig. \ref{fig:2}. This figure illustrates a trade-off between missing mask similarity and observed values correlation tuned by parameter $w$ as described in algorithm \ref{Missing-SCOP}. Notable improvement of clustering accuracy is observed in this case.

\begin{table}
\centering
\caption{Silhouette Values of each solution.}
\renewcommand{\arraystretch}{1.3}
\begin{tabular}{|c||c|c|c|}
	\hline
	\label{tab1}
	\backslashbox{Dataset}{Algorithm} & Impute   & non-Impute & no-Missing \\ \hline\hline
	$Cities$ 
	& 0.3802 & 0.3829 & 0.4221\\ \hline
	$Kmeansdata$ & 0.7958 & 0.8109 &   0.8606    \\ \hline
\end{tabular}
\end{table}

\section{Conclusion}
An innovative method of prediction enhancement is introduced and explained on linear models. SWP algorithm as a developed weighted least square solution is suggested and surpassed many state-of-the-art methods such as clustering in simulation results. Datasets containing missing information have been studied; adjusted SWP is developed for such scenarios, too. Clustering as a fundamental part of this adjustment is discussed and Missing-SCOP algorithm is introduced as a mean of handling missing values in clustering. Mentioned algorithm considers missing mask similarity of each example as a constraint of clustering by weight tuning parameter $w$. Comparing mean silhouette values as a factor of clustering precision, simulation results depicted that Missing-SCOP algorithm, a non-impute clustering method of cases with missing values, outperformed imputation methods like soft-impute.
\newline
\bibliographystyle{IEEEtran}
\bibliography{reffff.bib}
\end{document}